\documentclass[runningheads]{llncs}
\usepackage[T1]{fontenc}
\usepackage{graphicx}
\usepackage{amsmath,amssymb}
\usepackage{multirow}
\usepackage{wrapfig}
\usepackage{adjustbox}
\newcommand{\repeatthanks}{\textsuperscript{\thefootnote}}

\begin{document}
\title{Benchmarking Neural Decoding Backbones towards Enhanced On-edge iBCI Applications }
\titlerunning{Benchmarking Neural Decoding Backbones}
%
%

\author{Zhou Zhou\inst{1} \thanks{Zhou Zhou and Guohang He contribute equally to this work.}\and
Guohang He \inst{1} \repeatthanks \and
Zheng Zhang\inst{1,2} \thanks{Corresponding authors: Zheng Zhang and Ran Cheng.} \and Luziwei Leng\inst{2} \and Qinghai Guo\inst{2} \and Jianxing Liao\inst{2} \and Xuan Song\inst{1} \and Ran Cheng\inst{1}\repeatthanks}
\authorrunning{Z. Zhou and G. He et al.}
%
\institute{Southern
University of Science and Technology, Shenzhen 518055, China \email{ranchengcn@gmail.com}\and
Advanced Computing and Storage Lab, Huawei Technologies Co., Ltd.,
Shenzhen 518055, China \\ \email{zhangzheng147@huawei.com} }

\maketitle             
\begin{abstract}

Traditional invasive Brain-Computer Interfaces\,(iBCIs) typically depend on neural decoding processes conducted on workstations within laboratory settings, which prevents their everyday usage. Implementing these decoding processes on edge devices, such as the wearables, introduces considerable challenges related to computational demands, processing speed, and maintaining accuracy. This study seeks to identify an optimal neural decoding backbone that boasts robust performance and swift inference capabilities suitable for edge deployment. We executed a series of neural decoding experiments involving nonhuman primates engaged in random reaching tasks, evaluating four prospective models, Gated Recurrent Unit\,(GRU), Transformer, Receptance Weighted Key Value\,(RWKV), and Selective State Space model (Mamba), across several metrics: single-session decoding, multi-session decoding, new session fine-tuning, inference speed, calibration speed, and scalability.
The findings indicate that although the GRU model delivers sufficient accuracy, the RWKV and Mamba models are preferable due to their superior inference and calibration speeds. Additionally, RWKV and Mamba comply with the scaling law, demonstrating improved performance with larger data sets and increased model sizes, whereas GRU shows less pronounced scalability, and the Transformer model requires computational resources that scale prohibitively.
This paper presents a thorough comparative analysis of the four models in various scenarios. The results are pivotal in pinpointing an optimal backbone that can handle increasing data volumes and is viable for edge implementation. This analysis provides essential insights for ongoing research and practical applications in the field.
\end{abstract}
\section{Introduction}
Advancements in invasive Brain Computer Interfaces\,(iBCIs) have demonstrated promising results across various applications, including speech decoding~\cite{heelan2019decoding,willett2021high,wilson2020decoding}, prosthesis control~\cite{gilja2012high,xu2019unsupervised}, neurological disorders rehabilitation~\cite{cheng2020brain,lazarou2018eeg,stanslaski2012design,stanslaski2018chronically} and more. Accurate decoding the brain activities is crucial for the success of these applications. Previous efforts have focused on employing adaptive filters such as Kalman Filters~\cite{gilja2012high,wu2002neural,wu2008real} or traditional machine learning models such as Recurrent Neural Networks\,(RNNs)~\cite{sussillo2012recurrent,willett2021high}. However, with the expansion of the available neural data, significant progress has been made using Transformer-based architectures. Models such as Neural Data Transformer (NDT1)\cite{ye2021representation} leverage multi-session, multi-task and multi-subject neural data, yielding improved decoding performance and enhanced generalization capabilities with unseen data.

Limitations still exist among these methods. Despite the advantages of RNNs for handling long-term dependency, their inherent serial dependency significantly affect the model's inference speed~\cite{ye2021representation}. Meanwhile, it remains unclear whether scaling up GRU model size with data volume improves neural decoding accuracy. Transformers facilitate parallel computation and adhere to the scaling laws~\cite{DBLP:conf/emnlp/IvgiCB22}, but the increase in model size and sequence length leads to quadratic growth in model complexity (O($n^2$)), requiring a dramatic escalation in computational resources in order to fit in edge-device for portable BCI applications in daily use. 

Models such as Receptance Weighted Key Value model\,(RWKV)~\cite{DBLP:conf/emnlp/PengAAAABCCCDDG23} and Selective State Space model\,(Mamba)~\cite{gu2023mamba} have been designed utilizing linear attention mechanisms that offer reduced temporal and spatial complexity compared to traditional transformers. These models have demonstrated competitive performance in natural language processing and computer vision tasks~\cite{gu2023mamba,DBLP:journals/corr/abs-2401-10166,DBLP:conf/emnlp/PengAAAABCCCDDG23}, but it remains unclear which model is most suitable as the backbone for neural decoding. 

This paper investigates whether recent advancements in model architectures can enhance neural decoding. Instead of benchmarking against state-of-the-art\,(SoTA) architectures, we compare the RWKV and Mamba models with the GRU and Transformer models in terms of computational efficiency and decoding accuracy. We have designed a series of experiments to assess various parameters: decoding accuracy, adaptiveness to new sessions, inference time, and scalability trends on model size, to identify an optimal neural decoding backbone. To the best of our knowledge, this work might be the first effort to investigate linear attention mechanisms in neural decoding, targeting fast and low-power inference on edge devices.

\section{Related Work}
\subsection{Neural Decoding}

Neural decoding primarily relied on adaptive filters or traditional machine learning methods such as Kalman Filters~\cite{gilja2012high,wu2002neural,wu2008real}, Wiener Filters~\cite{hochberg2006neuronal} or SVM~\cite{taghizadeh2015decoding}. However, with the advent of deep learning, particularly the emergence of large-scale models, there has been a significant shift in neural decoding approaches. Deep learning models facilitate automated feature learning, reducing the impact of subjective factors and greatly improving decoding accuracy and efficiency. Recurrent neural networks and Transformers have now found more applications in neural decoding tasks~\cite{DBLP:journals/corr/abs-1901-05498}. Contemporary applications of brain decoding technologies extend to medical rehabilitation, assistive communication, and human-computer interaction~\cite{rapeaux2021implantable}. 

\subsection{RWKV}
Transformer has precipitated as disruptive revolution, particularly due to its widespread application of attention mechanisms across multiple domains. However, a significant issue arises as the memory and computational complexity of the Transformer grows quadratically with increasing sequence length. Concurrently, RNNs exhibit linear growth in memory and computational demands but are significantly outperformed by Transformer due to limitations in parallelization and scalability~\cite{DBLP:conf/emnlp/PengAAAABCCCDDG23}. To address this challenge, Bo Peng et at. have proposed the RWKV, which integrates the efficient parallel training advantages of Transformer with the effective inference comparable to that of similarly scaled Transformer, underscoring its potential and effectiveness in handling large-scale sequence data~\cite{DBLP:conf/icml/AlamRB0H23}.
\subsection{Mamba}
The state space model is a mathematical framework used to describe the evolution of systems over time. It employs state vectors to represent the current state of the system and uses state transition equations and observation equations to correlate the changes between system states and the relationship with observed data~\cite{DBLP:conf/iclr/GuGR22}. Mamba is an enhanced approach based on the structured state space model S4, integrating the recurrent structure of recurrent neural networks and the parallel characteristics of convolution neural networks. This approach excels in capturing long-term dependencies in sequential data and facilitates efficient parallel computation. By combining structured state space models with deep learning techniques, Mamba can handle sequential data more effectively, exhibiting higher modeling capability and predictive performance. Mamba has demonstrated superior performance in various domains, including language modeling, DNA sequence modeling, audio modeling and generation~\cite{gu2023mamba}.
\section{Methods}
\begin{figure}[t]
    \centering
    \vspace{-1.5em}
    \includegraphics[width=\textwidth]{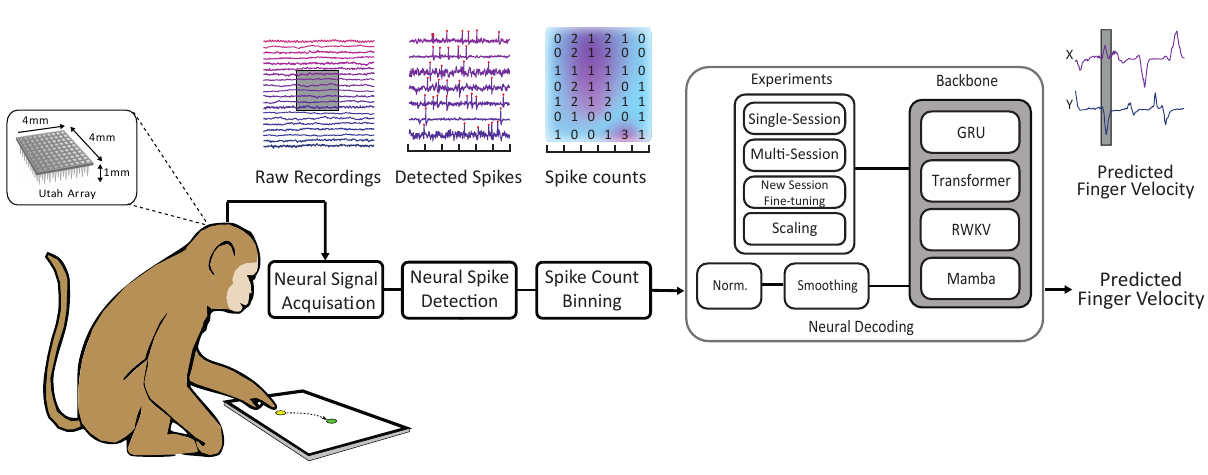}
    \caption{Raw neural signals were recorded from the primary motor cortex (M1) area of a monkey using a 96-channel Utah microelectrode array during random reach tasks. Spike activity detected from these neural signals was binned temporally across the 96 channels. The resulting matrix of spike counts served as inputs for various methods after normalization and smoothing, and the outputs were the predicted finger velocities along the x and y axes. Experiments conducted under different scenarios facilitated comparisons of predictive accuracy, inference speed, and scalability among the four types of backbone models.}
    \label{system}
    \vspace{-1.5em}
\end{figure}
The system architecture is shown in Fig.\ref{system}. The raw neural recording from the Utah array is processed into spike count bins and decoded using GRU, Transformer, RWKV and Mamba as four different backbones. The decoded output is compared with the ground truth motion activities. The detailed workflow is given below.
\subsection{Data Processing}
\subsubsection{Datasets} The dataset from~\cite{makin2018superior} is used in this study, which includes a rich collection of neural and behavioral data recorded from nonhuman primates engaged in a random target reaching task. This task requires the subject controlling a ticker to move the computer cursor and reach a series of randomly distributed targets displayed on screen in succession. During the execution of the task, the neural activities from primary motor cortex\,(M1) and primary sensory cortex\,(S1) are collected using Utah array, and the position of the subject's hand kinematic trajectories are recorded using motion tracking systems.

The neural recordings in this dataset consists of extracellular spike recordings, with the event times of threshold crossings sorted into discrete units. The recordings collected from subject Indy are used in this studies in total 30 sessions. The kinematic measurements contain the x and y coordinates of the subject's fingertip and cursor position as it reaches out, as well as the x and y coordinates of the set targets, both sampled at a frequency of 250 Hz. 

\subsubsection{Data processing}
In this studies, we only used recordings collected from the M1 cortex. We partitioned each session of the recorded task into multiple temporal bins with duration of 10\,ms. Due to the sampling rate of 250 Hz, the sampling frequency is increased to 1000 Hz using linear interpolation. Within these bins, we quantified the number of spike events\,(threshold crossings) for each neural recording channel, thereby capturing the discrete neural firing patterns over time. It is worth noting that we use the unsorted spike events known as multi-unit activities. In practice, spike sorting can be require too much computation for on-chip processing while using the sorted single-unit activities only bring limited decoding accuracy improvement as shown in ~\cite{todorova2014sort}. 

The cursor's velocity is used to characterize the kinematics of the reaching movement. The binned spike event and cursor velocity were temporally aligned, normalized and smoothed with a Gaussian smoothing operation, which attenuates high-frequency noise and elucidate the underlying signal trends following ~\cite{keshtkaran2022large}. 

In the experiments, the input to the model is denoted as $Spk \in \mathbb{R}^{S \times C}$, where $S$ represents the timesteps used for each prediction and $C$ denotes the number of channels. The ground truth denoted as $Vel \in \mathbb{R}^{S \times 2}$, which represents the finger speed in x and y axis at each timestep.

\subsection{Backbone Models} \label{models}

\subsubsection{GRU}
Proposed by Cho et at.~\cite{DBLP:journals/corr/BahdanauCB14}, GRU is a variant of the Recurrent Neural Network\,(RNN), specifically designed to address the challenges of gradient explosion and gradient vanishing in training. GRU achieves this by employing update gates and reset gates, which selectively update useful information and capture of long-term dependencies within time series data. GRU can be characterized by the following formulations~\cite{DBLP:journals/corr/ChungGCB14}:

\begin{align}
    h_t^j &= (1- z_t^j) \odot h_{t-1}^{j} + z_t^j \odot \tilde{h}_t^j \\
    z_t^j &= \sigma(W_z x_t + U_z h_{t-1}^j) \\
    \tilde{h}_t^j &= \tanh(W x_t + U(r_t \odot h_{t-1}^j)) \\
    r_t^j &= \sigma(W_r x_t + U_r h_{t-1}^j)
\end{align}

The reset gate\,($r$) is a gating mechanism that modulates the flow of information from the previous activation, allowing the model to discard irrelevant past state information, thus mitigating the vanishing gradient problem. The update gate\,($z$) determines the extent to which the unit updates its activation, or hidden state\,($h$). It controls the degree of information transfer from the previous state to the current state, enabling the model to capture long-term dependencies. The activation\,($h$), commonly referred to as the hidden state, captures the learned information at the current time step and is recursively influenced by past activation. In our work, we employ hidden size $d_h = 256$. The candidate activation\,($\tilde{h}$) is a proposed update to the hidden state, which incorporates new input while being modulated by the reset gate to potentially discard the irrelevant previous state.
\subsubsection{Transformer}
The foundational mechanism of the Transformer is its self-attention mechanism, which enables the model to dynamically adjust the weighting of input data, such as tokens or sequence elements, based on their contextual relevance~\cite{DBLP:conf/nips/VaswaniSPUJGKP17}. Unlike GRUs, which process data sequentially, Transformers handle input in parallel during the training phase, significantly expediting the training process. 

We employ the classic Multihead Scaled Dot-Product Attention mechanism along with an encoder-decoder architecture. Unlike traditional approaches that transform input vectors of vocabulary tokens through an embedding layer to embed feature dimensions, we directly take the spike matrix \(x \in \mathbb{R}^{S \times C}\)  as the input for both encoder and decoder and treat the channel dimension $C$ of the input spike matrix as the feature dimension and project the feature dimension to the model hidden dimension following Eq.\ref{tsf-input} and this projection is also used in RWKV and Mamba model.

\begin{equation}
A = f(x) = Wx + b \label{tsf-input}
\end{equation}
\begin{equation}
B = E[\text{positions}]
\end{equation}
\begin{equation}
input = \text{Dropout}(A + B)
\end{equation}
\begin{equation}
output = \text{Decoder}(\text{Encoder}(input), input)
\end{equation}

The function $f$ represents a linear mapping layer, where $W$ and $b$ denote the weights and biases of the input layer, respectively. 
$E$ corresponds to the positional embedding matrix, from which an embedding vector is selected for each positional index. Where \(input \in \mathbb{R}^{S \times d_{\text{model}}}\) is the input to the encoder and decoder and \(output \in \mathbb{R}^{S \times 2}\) is the predicted x and y axis velocity. In the encoder and decoder, the attention is implemented as below:

\begin{align}
\text{Attention}(Q, K, V) &= \text{softmax}(\frac{QK^T}{\sqrt{d_k}})V \label{attn}
\end{align}
\begin{gather}
\begin{aligned}
\text{MultiHead}(Q, K, V) &= \text{Concat}(\text{head}_1, \ldots, \text{head}_h) W^O \\
\text{where } \text{head}_i &= \text{Attention}(Q', K', V')
\end{aligned}
\end{gather}

Where the $Q, K, V$ are calculated following Eq.\ref{tsf-input} with independent weights and zero bias, the parameter matrix \(W^O \in \mathbb{R}^{hC \times d_{\text{model}}}\) and \(Q' \in \mathbb{R}^{S \times h \times d_q}\), \(K' \in \mathbb{R}^{S \times h \times d_k}\), \(V' \in \mathbb{R}^{S \times h \times d_v}\). Here, we employ $h$ = 2 heads, \(d_q = d_k = d_v = \frac{d_{\text{model}}}{h}\) and $d_{model} = 128$.

Given the limited variance in input data patterns, the data is processed through two separate attention heads. The system comprises three layers each of encoders and decoders, culminating in the prediction of velocities in the x and y axes.

\subsubsection{RWKV}
Unlike most RNNs, RWKV is a recurrent model combines the efficient parallelizable training of transformers with the fast inference time. RWKV reformulates the attention mechanism with a variant of linear attention, replacing traditional dot-product token interaction with more effective channel-directed attention~\cite{DBLP:conf/emnlp/PengAAAABCCCDDG23}.
It mitigates the memory bottleneck and quadratic scaling issues inherent in Transformers through efficient linear scaling. It also preserves the ability for parallelized training and ensures robust scalability.

\begin{align}
    r_t &= W_r(\mu_r\odot x_t + (1-\mu_r)\odot x_{t-1})\\
    k_t &= W_k(\mu_k\odot x_t + (1-\mu_k)\odot x_{t-1})\\
    v_t &= W_v(\mu_v\odot x_t + (1-\mu_v)\odot x_{t-1})
\end{align}

R encodes historical information, activated via a Sigmoid function and incorporating a forgetting mechanism. W signifies the positional weight decay vector, a trainable parameter within the model. The terms 
K and V function analogously to the key and value in Transformer architectures. Distinct from traditional models where $x$ is simply the embedding of the current token, in the RWKV, 
$x$ is calculated as the weighted sum of the embeddings of the current token and the previous token.
\begin{equation}
wkv_t = \frac{\sum_{i=1}^{t-1} e^{-(t-1-i)w+k_i} \odot v_i + e^{u+k_t} \odot v_t}{\sum_{i=1}^{t-1} e^{-(t-1-i)w+k_i} + e^{u+k_t}}
\label{wkv_t}
\end{equation}
Equation \ref{wkv_t} functions similarly to an attention mechanism, representing position $t$ as a learnable weighted sum of past content. In RWKV, $w$ is treated as a channel-wise vector that adjusts according to the relative position, requiring the training of only a single parameter vector $w$. $u$ is designated for individual processing of the current token's position, serving to circumvent any potential degradation of $w$. 
\subsubsection{Mamba}
In contrast to the quadratic scaling observed with traditional models, Mamba demonstrates a throughput up to five times faster than the Transformer and exhibits linear scaling with sequence length~\cite{gu2023mamba}. Unlike RNNs, which compress all information into a hidden space and struggle with long-term memory issues, Mamba introduces a selective state-space model. This model offers the benefits of a linear recurrent network, enhanced by mechanisms for rapid training and effective context retention. Improvements in Mamba’s Structured State Spaces\,(SSM) include a selection mechanism that filters out irrelevant information while enabling indefinite memory retention, and a hardware-aware algorithm optimized for GPU memory layouts to facilitate hardware acceleration. This ensures efficient computation cycling without extending the state unnecessarily, thus enhancing performance.

The SSM Mamba consists of the following two equations:
\begin{equation}
    x_t=f(x_{t-1}, u_t, w_t) \label{state transition equation}
\end{equation}
\begin{equation}
    y_t=h(x_t, v_t) \label{observation equation}
\end{equation}
Equation \ref{state transition equation} represents the state transition equation, describing how the system state evolves over time. Here, $x_t$ denotes the system state at time step $t$, $u_t$ represents the control input, $w_t$ is the process noise, and f is the state transition function. Equation \ref{observation equation} is the observation equation, $y_t$ represents the observation data at time step $t$, $v_t$ denotes the observation noise, and $h$ is the observation function. The concept of selectivity in Mamba allows the model to selectively remember or forget information at each time step.

\section{Experiments and Key Results}
\subsection{Experiment settings}
To evaluate the capabilities of different backbone models across various dimensions, four distinct experiments were established: single-session, multi-session, new session finetuning, and scaling experiments\,(set timestpes as 128, 1024, 128 and 1024 respectively). A total of 30 sessions, collected over different days from the same subject, were used. All neural recordings from these 30 days were divided into training and testing datasets with an 8:2 ratio, consistently applied across all experiments.

Single-Session Experiment: This experiment assessed the ability of the backbone models to perform effectively on small datasets. Each of the four models was trained independently on data from individual sessions, with recording lengths varying from 360\,s to 3363\,s.

Multi-Session Experiment: This experiment focused on the models' capacity to extract deep latent representations from neural recordings with input feature shifting overtime. A unified model was trained using training sets from all sessions. Over time, the quality of the recordings degraded due to scar tissue encapsulation around the implants, leading to increased noise levels and a decrease in detected neural firing rates from over 20Hz to below 10Hz. Additionally, the neurons observed on different channels changed over time. Various training strategies were explored to help models adapt to these shifting input features.

New Session Finetuning Experiment: This experiment tested the models' ability to generalize and adapt to unseen data. Models were initially trained with datasets from the first 25 days, and then incrementally finetuned using datasets from the last five days (10 seconds per iteration). This setup mirrors practical scenarios for BCI calibration on new days, where a shorter calibration time is often critical. The aim was to identify the model that could quickly return to acceptable performance levels, making it more suitable for real-world use outside the laboratory.

Scaling Experiment: This experiment investigated whether increased model size could enhance performance. The scaling law has been a key principle in designing large language models~\cite{DBLP:journals/corr/abs-2001-08361}, but its applicability in neural decoding remains unexplored.

\begin{table}
\centering
\vspace{-1.5em}
\caption{Parameter counts and hyperparameters of models}\label{tab1}
\begin{adjustbox}{width=\textwidth}
\begin{tabular}{ccccc}
\hline
Model       & Parameters &  Epochs Single(Multi)   & Layers & Embedding Size\\ 
\hline
GRU         & 272k     & 30(50)    & 1    &256     \\
Transformer & 316k     & 50(50)    & 3    &128     \\
RWKV        & 294k     & 30(50)    & 2    &88     \\ 
Mamba       & 306k     & 30(50)    & 2    &144     \\ \hline
\end{tabular}
\end{adjustbox}
\vspace{-1.5em}
\end{table}

The same hyperparamter settings are used for all experiments except the scaling experiment, with details on their parameter counts and hyperparameters presented in Table.\ref{tab1}. The requirement for the Transformer model to undergo 50 epochs may be attributed to its attention mechanism, which necessitates numerous iterations to effectively optimize attention weights. Additionally, the design of the Transformer, which processes entire sequences simultaneously, may contribute to slower convergence rates during training.~\cite{DBLP:conf/nips/VaswaniSPUJGKP17}.

The R$^2$ is used to evaluate the neural decoding performance following~\cite{ahmadi2021robust,zhang2023firing}. R$^2$ typically ranging from 0 to 1, an R-square value of 0 indicates that the model fails to explain any variance in the dependent variable, while a value of 1 indicates a perfect fit if the model to the data. The formula for calculating R$^2$ is as follows:
\begin{align}
    RSS &= \sum_{i=1}^{n}(y_i-\hat{y}_i)^2 \\
    TSS &= \sum_{i=1}^{n}(y_i-\bar{y})^2
\end{align}
\begin{equation}
    R^2=1-\frac{RSS}{TSS}
\end{equation}
where $RSS$ is the residual sum of squares(the sum of the squares of the differences between actual($y_i$) and predicted values($\hat{y}_i$)), and $TSS$ is the total sum of squares(the sum of the squares of the differences between actual values and the mean of the observed values($\bar{y}$)). Table.\ref{tab2} summarizes the evaluation results of four models in different experiments. 

\subsection{Single-session experiment}

As shown in Table.~\ref{tab2}, the RWKV model excels in the single-session experiment, surpassing the GRU model by 0.02 in R$^2$. However, both the Mamba and Transformer models score below 0.7, indicating that these models are less effective when dataset sizes are limited.

In terms of inference time processing 1280\,ms of neural data\,(1 batch), as detailed in Table.~\ref{tab2}, shows varying performance among the models. The GRU model requires the longest processing time due to its sequential processing nature. The Transformer model also exhibits relatively long inference times due to its computationally intensive operations. In contrast, the RWKV and Mamba models demonstrate significant advantages in inference speed over both the GRU and Transformer models.

Specifically, RWKV, which is a recurrent neural network devoid of an attention mechanism, avoids the computational overhead associated with computing attention matrices. This model incorporates Token Shift and Channel Mix mechanisms to optimize position encoding and channel blending, thereby enhancing both efficiency and performance. On the other hand, Mamba achieves rapid inference and maintains linear scalability with sequence length through dynamic and selective retention or dismissal of information based on input. Its streamlined and homogeneous architecture, coupled with a selective state space, markedly boosts inference speed.
\subsection{Multi-session experiment}
In the multi-session experiment, we explored three different data partitioning strategies during training to identify the most effective approach for aiding models to learn as input features shifted. These strategies are as follows:
\begin{itemize}
\item[$\bullet$] Random partitioning: Batches are randomly selected from random sessions to be fed into the model.
\item[$\bullet$] Sequential partitioning: Data batches are fed into the model in a sequential order, day by day.
\item[$\bullet$] Random session partitioning: Sessions are selected randomly, but within each selected session, data batches are fed sequentially.
\end{itemize}

The random training strategy results in significantly higher stability and decoding accuracy of the model compared to the other two strategies. Although the data are strongly time-correlated, this approach of random input enhances gradient diversity, reduces cyclic biases in data appearance, and helps prevent overfitting.

Sequential training resulted in limited improvement over the single-session experiment for both the GRU and RWKV models. Although these models can memorize historical information, sequential training may still lead to catastrophic forgetting, thereby only marginally enhancing performance compared to the single-session results. In contrast, the Mamba model demonstrated a significant improvement, nearly 0.1 increase in R$^2$, over the single-session experiment. This suggests that Mamba's selective state space mechanism is more effective at preserving useful information and handling long-term dependencies compared to the gating mechanisms of GRU or the RWKV model in neural decoding.

However, the random session training strategy failed to provide a diverse training gradient and the data order could not convey long-term dependencies, resulting in underfitting of the model.

Another observation during the multi-session training is the difficulty in achieving convergence with the Transformer model, which required careful tuning of its hyper-parameters. In contrast, the other models exhibited less sensitivity to training hyper-parameter settings.
\begin{table}[t]
\centering
\caption{Experiments results on all models} \label{tab2}
\begin{adjustbox}{width=\textwidth}
\begin{tabular}{c||ccccc}
\hline
\multicolumn{1}{l}{Experiment}  & Indicator                  & GRU   & Transformer & RWKV  & Mamba \\ \hline
\multirow{2}{*}{Single-session} & Average R$^2$        & 0.715 & 0.633       & \textbf{0.717} & 0.660 \\
                                & Inference time/s  & 0.941 & 0.822       & \textbf{0.303} & 0.434 \\ \hline
\multirow{3}{*}{Multi-session}   & \textbf{Random train}   & \textbf{0.838} & 0.720       & 0.812 & 0.810 \\
                                & Sequence train & 0.749 & 0.523       & 0.726 & \textbf{0.752} \\
                                & Random session & 0.560 & 0.314       & 0.600 & 0.556 \\ \hline
\multirow{2}{*}{Fine-tuning}       & Average R$^2$        & \textbf{0.773} & 0.748       & 0.763 & 0.756 \\
                                & Recovery time/s   & 214   & -         & 202   & \textbf{178}   \\
                                & Zero shot         & 0.4811	& 0.383	& 0.452 & 0.370 \\ \hline
\multirow{2}{*}{Scaling}        & Max R$^2$         & 0.846 & -           & 0.843 & \textbf{0.851} \\
                                & Increment         & 0.010  & -           & 0.031 & \textbf{0.041}      \\ \hline
\end{tabular}
\end{adjustbox}
\vspace{-1.5em}
\end{table}

\subsection{Fine-tuning the model on new sessions}

As shown in Table.\ref{tab2}, among the four models, GRU achieved the highest average R$^2$ score over 5 days of fine-tuning on new sessions, reaching 0.773. The RWKV and Mamba models scored 1-2\% lower, while the Transformer model recorded the lowest score at 0.748. Regarding zero-shot performance, we only saw RWKV achieved an R$^2$ of 0.7 in one session out of five. On average, none of the models achieved adequate zero-shot performance.

The results from the finetuning experiment indicate that all models are capable of surpassing their performance when trained solely on single-session data. This demonstrates that despite variations in firing rates and neuron-channel mappings over time, the models can distill useful information to enhance neural decoding. The quality of the base model significantly influences the effectiveness of the finetuned model. However, the backbone model alone does not provide zero-shot capability, suggesting that additional architectural designs or training strategies are necessary to enhance the models' adaptability to input feature shifts and improve zero-shot performance.

In terms of the data length required to achieve an acceptable R$^2$ score of 0.7 through fine-tuning, Mamba outperformed both RWKV and GRU. This superior performance likely stems from Mamba’s enhanced ability to resolve long-term dependencies, which facilitates its calibration to unseen data more effectively. Consequently, Mamba emerges as a more viable option for real-world deployment in practical BCI applications due to its robust adaptability.

\subsection{Scaling analysis}

\begin{wrapfigure}{r}{7cm}
  \includegraphics[width=7cm]{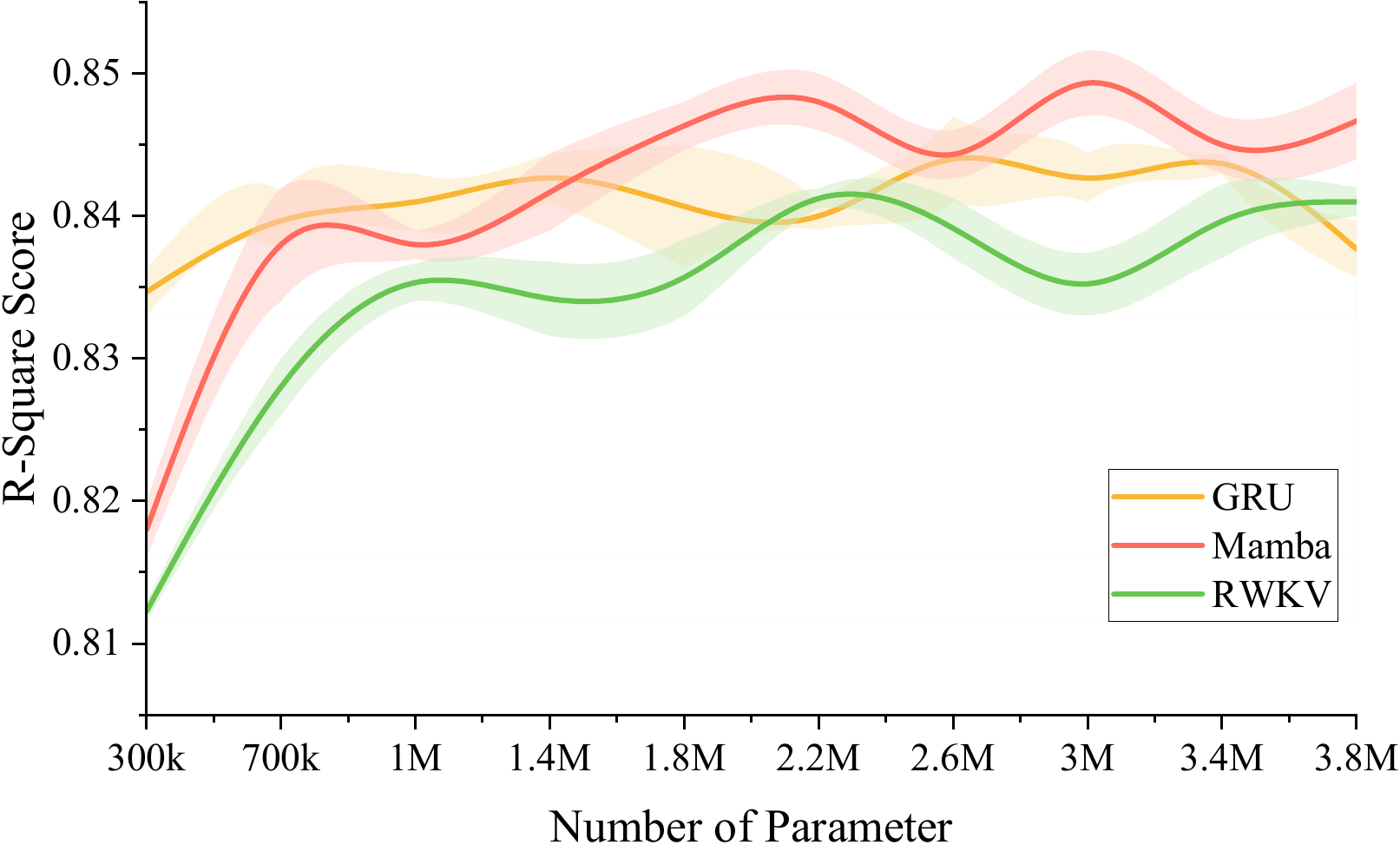}
  \caption{Scaling parameter counts for the models range from 300k to 3.8M with error}
  \label{scaling}
  \vspace{-1.5em}
\end{wrapfigure}

In multi-session training, the parameter count for the models we used is approximately 300k. To explore whether increasing the model size could enhance its decoding performance, we examined the improvements achieved by increase the number of layers in GRU, RWKV and Mamba models\,(Transformer can fail to converge in many cases and is therefore ignored.). The variation in the model's decoding R$^2$ scores as a function of the parameter count of these models, ranging from 300K to 3M, is illustrated in the Fig.\ref{scaling}.

With an increase in model parameters, the R$^2$ scores for Mamba and RWKV show significant improvement, reaching 0.843 and 0.851 respectively. This represents increases of 0.031 and 0.041 over their 300k parameter models. In contrast, the GRU model demonstrates only a mild improvement of 0.01 when parameters are increased, and further scaling leads to a declining trend in performance. Despite its gate mechanisms to mitigate the vanishing gradient problem, GRU’s inherent sequential processing nature restricts its scalability and limits its efficiency in handling large-scale sequence tasks.

Conversely, RWKV and Mamba exhibit superior scalability and computational efficiency, outperforming GRU. This advantage is largely due to their innovative structural designs and optimization strategies that effectively address the limitations typically associated with recurrent neural networks and traditional Transformers.

While performance gains for RWKV and Mamba level off as model size increases, this plateau is mainly attributable to the limited size of the current dataset. However, with the rapid advancement of BCI technology and the anticipated increase in available data, it is reasonable to predict that RWKV or Mamba could serve as robust backbones for neural decoding in future applications.

\section{Discussion}

\subsection{Suggestions on model selection}

Each of the four models evaluated demonstrates distinct strengths and weaknesses. The GRU model achieves the secondary prediction accuracy in single session experiment and best predictive accuracy on multi-session experiment on the dataset used in this work. However, its inference time and calibration recovery time is constrained by its inherent serial structure. In contrast, the RWKV and Mamba model have significantly faster inference and calibration recovery time. Additionally, both RWKV and Mamba adhere to the scaling law, demonstrating a gradual improvement in predictive accuracy as model sizes increase. Mamba eventually achieves an R$^2$ of 0.851 when scale up to 3M, hitting the highest score among all models in different experiment settings. It also becomes compatible with the SoTA neural decoding model POYO~\cite{DBLP:conf/nips/AzabouAGMNMRPLD23}, trained on a much larger dataset tested on the same task. The Transformer model, however, lags in nearly all performance metrics and is difficult to converge in our experiments.

Consequently, Mamba or RWKV could be suitable backbones for future neural decoding tasks, especially with an increasing amount of available neural recordings. Their scalability and linear computational complexity can significantly enhance decoding performance without the need for excessive computational resources, making them preferable for wearable devices used daily. For BCI applications, this choice can also lead to reduced training times, faster response times, and quicker calibration speeds. However, for studies involving a limited amount of data and those not sensitive to response times, RNN models like GRU or LSTM may suffice to provide high decoding performance in most use cases.

\subsection{Limitation and future works}

One significant challenge within the BCI field is achieving long-term stable neural decoding. Unfortunately, none of the four models can provide long-term stable decoding capabilities without finetuning, based on our experiments. While this work utilizes only one dataset, introducing a more diverse dataset could enable the model to learn a broader array of data features, thereby enhancing its robustness in practical applications.

The degradation in long-term decoding performance is primarily due to input feature shifting~\cite{zhang2023firing}. To manage potential data drift over prolonged periods, continuous or online learning strategies could be implemented, allowing the model to continually adapt to new data. From a computational-saving perspective, instead of full parameter updating, tuning only the input and output layers or employing some transfer learning strategies might better accommodate input variations with less computational overhead.

New training strategies can also be explored to guide the model in learning useful latent representations. By implementing a weighted loss scheme that prioritizes recent sessions chronologically, our preliminary results have already shown notably improved zero-shot outcomes.

Additionally, the backbone models in this study were only trained on a random track task with one subject. The adaptation across different tasks and subjects also needs to be carefully evaluated in future studies.

\section{Conclusion}

This study has conducted a comprehensive comparison of GRU, Transformer, RWKV, and Mamba models in the context of neural decoding for random reach tasks. RWKV and Mamba, which demonstrate faster inference speeds, lower computational complexity, better scalability compared to GRU and Transformer, emerge as preferred choices for deployment on wearable devices. This detailed evaluation of the various strengths and weaknesses of each model not only highlights their individual capabilities but also establishes a robust foundation for future advancement on model architecture. The insights gained from this work guide the development of more efficient and effective neural decoding architecture, paving the way for enhanced performance in practical applications.

\section{Acknowledgement}
This work was supported in part by Guangdong Natural Science Funds for Distinguished Young Scholar under Grant 2024B1515020019.
%
%
%
\bibliographystyle{splncs04}
\bibliography{main}
%





\end{document}